\documentclass[lettersize,journal]{IEEEtran}
\usepackage{amsmath,amsfonts}
\usepackage{algorithmic}
\usepackage{algorithm}
\usepackage{array}
\usepackage[caption=false,font=normalsize,labelfont=sf,textfont=sf]{subfig}
\usepackage{textcomp}
\usepackage{stfloats}
\usepackage{url}
\usepackage{verbatim}
\usepackage{graphicx}
\usepackage{cite}
\usepackage{array}
\usepackage{tabularx}
\usepackage{multirow}
\usepackage{booktabs}
\usepackage{enumitem}
\begin{document}

\title{Computer-Vision-Enabled Worker Video Analysis for Motion Amount Quantification}

\author{Hari~Iyer,
        Neel~Macwan,
        Shenghan~Guo,
        and~Heejin~Jeong
\thanks{The corresponding author, Heejin Jeong, is from The Polytechnic School at Arizona State University, Mesa, AZ, 85212, USA.
E-mail: heejin.jeong@asu.edu}% <-this % stops a space
\thanks{The first author, Hari Iyer, is from The Polytechnic School at Arizona State University, Mesa, AZ, 85212, USA.}% <-this % stops a space
\thanks{Neel Macwan and Shenghan Guo are from The School of Manufacturing Systems and Networks at Arizona State University, Mesa, AZ, 85212}% <-this % stops a space
\thanks{Manuscript received xxxxxxx, 2024; revised xxxxxxx, 2024}}

\markboth{}%
{Shell \MakeLowercase{\textit{et al.}}: A Sample Article Using IEEEtran.cls for IEEE Journals}


\maketitle

\begin{abstract}
The performance of physical workers is significantly influenced by the extent of their motions. However, monitoring and assessing these motions remains a challenge. Recent advancements have enabled in-situ video analysis for real-time observation of worker behaviors. This paper introduces a novel framework for tracking and quantifying upper and lower limb motions, issuing alerts when critical thresholds are reached. Using joint position data from posture estimation, the framework employs Hotelling’s $T^2$ statistic to quantify and monitor motion amounts. A significant positive correlation was noted between motion warnings and the overall NASA Task Load Index (TLX) workload rating (\textit{r} = 0.218, \textit{p} = 0.0024). A supervised Random Forest model trained on the collected motion data was benchmarked against multiple datasets including UCF Sports Action and UCF50, and was found to effectively generalize across environments, identifying ergonomic risk patterns with accuracies up to 94\%.
\end{abstract}


\section{Introduction}
\IEEEPARstart{W}{orker} motion analysis involves studying human movement within workplace environments. Sustained awkward postures and muscular overexertion often lead to work-related injuries \cite{yang2017motion}. The primary motivation for analyzing worker motion is safety concerns. By monitoring worker movements, safety professionals and ergonomists can detect potential risks, such as injuries caused by repetitive muscular strain and awkward postures \cite{palikhe2020analysis}. Implementing safety measures and suggesting ergonomic improvements can help mitigate risk of injuries.

Analyzing task execution in labor-intensive fields like manufacturing, construction, and agriculture can uncover redundant or wasteful movements. Gouett \cite{gouett2010activity} analyzed activity for continuous productivity improvement in construction and found that applying this process continually to a construction site can significantly improve direct-work rates throughout a project's duration. This optimization not only increases productivity but also reduces unnecessary strain on workers. Worker motion analysis is closely associated with ergonomics \cite{bortolini2020motion}, which is the science of designing workplace tools and tasks to fit the worker. For example, Golabchi et al. \cite{golabchi2015automated} developed a programmed simulation of biomechanical systems for design analysis of workplace ergonomics. This approach lowers ergonomic risks by detecting uncomfortable postures in virtual models. Understanding how workers interact with products and equipment allows designers to create tools that are easier and more comfortable to use.

Worker motion analysis supports training and skill development. Organizations can develop motion simulation-based programs to train new employees efficiently \cite{macwanhighfidelity}. Motion analysis can identify areas where workers may require additional support or training. A critical aspect of worker motion analysis is human data collection capability. Recently, with the advances in sensing and artificial intelligence (AI) technologies, real-time sensing has enabled task-based data collection \cite{5753937}. For example, the posture- and position-based workers’ safety risk evaluation framework developed by Chen et al. \cite{chen2019proactive} demonstrates that automated evaluation of safety risk extent, combining spatial and postural factors, is almost 85\% accurate, outperforming the approximately 80\% accuracy of single-feature-driven risk examination using location, and 55\% accuracy when using posture. This proactive approach mitigates the risk of musculoskeletal injuries. According to Ryu et al. \cite{ryu2018motion}, motion data can be collected automatically to assess worker posture across experience levels. This postural analysis identifies areas where workers need training to improve their performance levels.

\begin{figure*}[h]
  \centering
  \includegraphics[width=\textwidth]{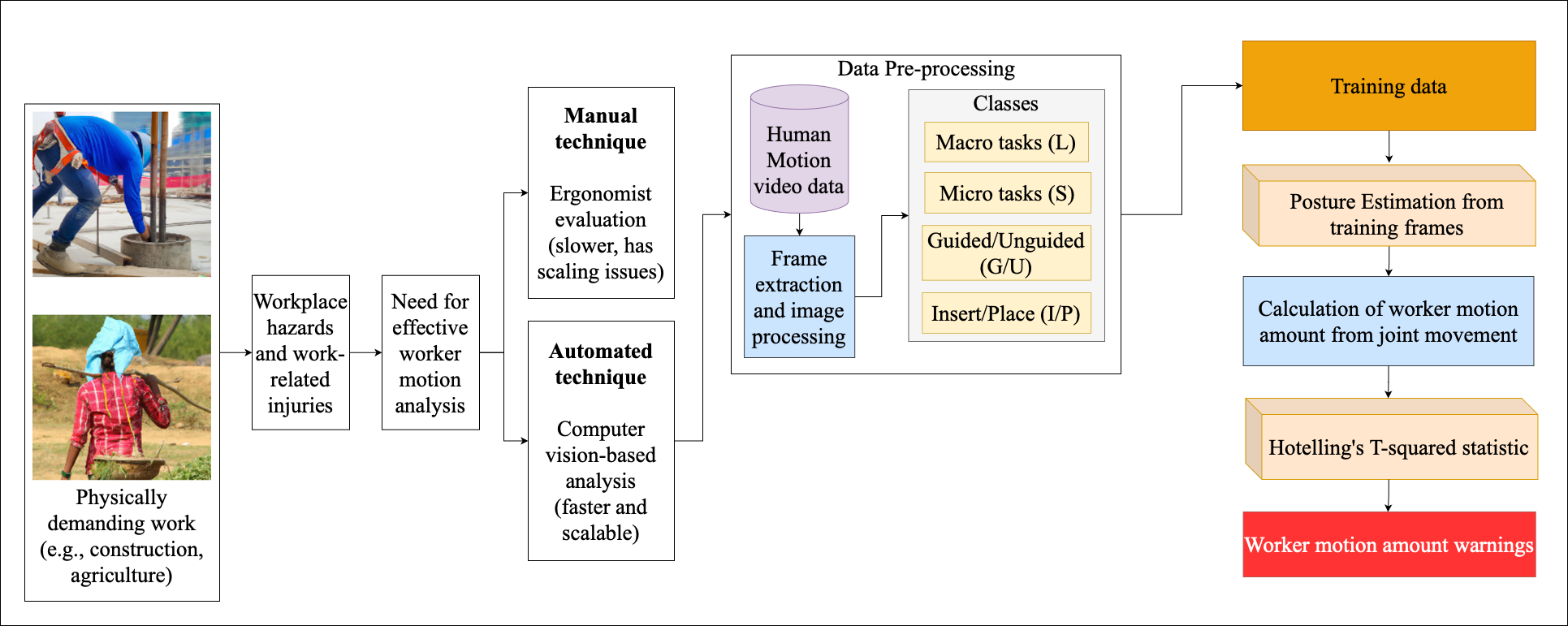}
    \caption{Architecture pipelines of posture estimation and worker motion analysis from the videos captured of assembly tasks.}
    \label{fig:Figure 1}
\end{figure*}

Current techniques for processing and analyzing human motion data are inefficient, making the monitoring and evaluation of joint movements challenging. Moran \& Wallace \cite{moran2007eccentric} reviewed studies involving vertical jumps and noted findings that were contradictory, suggesting inconsistent levels of eccentric loading and range of joint motion as potential causes. Video-based frame analysis is an improvement over traditional methods for studying human motion. For real-time processing of image frames, computer vision (CV) techniques use pose estimation and machine learning to extract actionable information \cite{10742356}. Mehrizi et al. \cite{mehrizi2018computer} estimated 3D posture in symmetrical lifting tasks, calculating kinematics using joint motion using CV-based motion capture compared against a surface marker-based system. Their results inferred that the CV-based approach aligned closely with the marker-based method. Seo et al. \cite{seo2016automated} developed a CV model for posture classification to perform automated ergonomic assessment, achieving over 90\% accuracy. Various methods are selected based on task and industry demands. In-situ videos provide real-time information about ergonomic behavior, but scalability in posture extraction and processing remains challenging.

\begin{table}[ht]
\centering
\caption{Notation Used in Motion Quantification and Risk Detection Framework}
\begin{tabular}{|l|p{4.5cm}|l|}
\hline
\textbf{Symbol} & \textbf{Definition} & \textbf{Dimensions} \\ \hline
\( \boldsymbol{v}_{i,k} \) & 3D position of landmark \( i \) at frame \( k \) & \(\mathbb{R}^3\) \\ \hline
\( \boldsymbol{m}_{i,k} \) & Motion vector of landmark \( i \) in frames & \(\mathbb{R}^3\) \\ \hline
\( \boldsymbol{m}_k \) & Concatenated motion vector \( k \) & \(\mathbb{R}^{3p}\) \\ \hline
\( M_k \) & Total motion magnitude in frame \( k \) & Unitless \\ \hline
\( \overline{M} \) & Mean motion across all frames & Unitless \\ \hline
\( \text{RMSD}_{\text{task}} \) & RMSD for task segment & Unitless \\ \hline
\( \boldsymbol{\mathcal{V}} \) & Position matrix for landmarks in frames & \( K \times 3p \) matrix \\ \hline
\( \boldsymbol{S} \) & Covariance matrix of motion vectors & \( \mathbb{R}^{3p \times 3p} \) \\ \hline
\( t_k \) & Hotelling’s \( T^2 \) statistic for frame \( k \) & Unitless \\ \hline
\( \boldsymbol{T} \) & Vector of \( T^2 \) values across frames & \(\mathbb{R}^K\) \\ \hline
\( \text{UCL} \) & Upper control limit from F-distribution & Unitless \\ \hline
\( \text{LCL} \) & Lower control limit (typically 0) & Unitless \\ \hline
\( y_i \) & True warning label for segment \( i \) & \{0, 1\} \\ \hline
\( \hat{y}_i \) & Predicted warning label for segment \( i \) & \{0, 1\} \\ \hline
\( \tau_d \) & Median warning threshold for dataset \( d \) & Integer (count) \\ \hline
\( \mathbf{x}_i \) & Feature vector: RMSD, mean, SD, UCL & \( \mathbb{R}^4 \) \\ \hline
\( f_\theta \) & Trained Random Forest classifier & \( \mathbb{R}^4 \rightarrow \{0, 1\} \) \\ \hline
\( h_j \) & Decision tree \( j \) in the forest & \( \mathbb{R}^4 \rightarrow \{0, 1\} \) \\ \hline
\( \mu_{\text{motion}} \) & Mean joint motion for a segment & Unitless \\ \hline
\( \sigma_{\text{motion}} \) & Standard deviation of joint motion & Unitless \\ \hline
\end{tabular}
\end{table}

To address these limitations, we propose an automated framework (see Fig. 1). The framework uses CV techniques \cite{bazarevsky2020blazepose} to extract body features from task videos through posture estimation. MediaPipe identifies 33 body landmarks; we focused on key landmark points for motion quantification \cite{lugaresi2019mediapipe}. Hotelling's $T^2$ control charts were built to flag significant “anomalies” \cite{montgomery2019introduction}. A correlation analysis was done between the detected anomalies and ground-truth motion information validated motion detection accuracy. We further implement a Random Forest classification model trained on motion statistics and warnings to assess generalization across diverse task domains. The proposed framework can be implemented in real-time to quantify and monitor worker motions. This study contributes to the field of instrumentation and measurement by transforming in-situ video data into quantifiable joint motion metrics using posture estimation pipelines and multivariate statistical monitoring. The proposed framework uses CV to collect data, calculates joint movement, and applies Hotelling’s \( T^2 \) method to detect motion anomalies, for real-time ergonomic assessment and task monitoring. The key contributions of this work are:
\begin{enumerate}[leftmargin=*]
\item We present a novel, video-based framework for ergonomic risk quantification using motion magnitude and multivariate monitoring of body landmark trajectories.
\item We implement a Hotelling’s \( T^2 \)-based control chart methodology to detect deviations in motion patterns and define warning thresholds for ergonomic risk.
\item We conduct an ablation study comparing full-body and partial landmark subsets (hands, upper body, lower body) across five benchmark datasets.
\item We introduce an adaptive thresholding mechanism using dataset-specific medians to improve classification of motion warnings, achieving high generalization from a single training source.
\item We demonstrate the effectiveness of our method through comparative classification performance, highlighting robustness across varied physical task domains and dataset types.
\item We incorporate multiple benchmark datasets, including G-AI-HMS \cite{iyer2025gaihms}, UCF Sports Action \cite{soomro2015action, rodriguez2008action}, UCF50 \cite{reddy2013recognizing}, and PE-USGC \cite{10742356}.
\end{enumerate}

The remainder of this paper is organized as follows. Section 2 reviews state-of-the-art literature on worker behavioral analysis and data science methods in this field. Section 3 describes the experimental designs and data collection process. Section 4 details the technical aspects of the proposed framework. Section 5 presents the results, and Section 6 discusses the findings. Finally, Section 7 concludes the paper with a summary and future directions.

\section{Literature Review}

\subsection{Hierarchical Approaches to Learning Human Behavior}
Recognizing human activities in videos is a challenging task. Robertson \& Reid \cite{robertson2006general} proposed a hierarchical approach combining trajectory information and local motion descriptors. However, this method requires a large database and accurate motion evaluation, making it less practical. Bregler \cite{bregler1997learning} used Expectation Maximization clustering, dynamical systems, and Hidden Markov Models to learn human dynamics in videos. Duchenne et al. \cite{duchenne2009automatic} introduced a weakly-supervised learning algorithm that extracts features from videos using scripts for weak supervision and clusters them based on feature similarity. A temporal action detector is then trained on these clusters. Han \& Bhanu \cite{han2007fusion} used a hierarchical genetic algorithm to find correspondence between colored and thermal images. Hoai et al. \cite{hoai2011joint} introduced a framework for video segmentation and action recognition using multi-class SVM and dynamic programming. Human video analysis faces a significant challenge due to rapid movements of humans.

In our case, we use CV to extract feature landmarks in videos, treating it as a pose estimation problem.

\subsection{CV Approaches for Human Task Video Analysis}
With the emergence of deep learning and GPU Computing, numerous architectures provide the flexibility to design networks tailored to specific tasks. Particularly in domains like human action recognition or behavior understanding, various models achieve high performance \cite{huang2021shallow, maswadi2021human}. Ji et al. \cite{ji20123d} used a 3D Convolutional Neural Network. This method extracts spatial and temporal features, but it has a high computational cost. Wang et al. \cite{wang2018temporal} presented a Temporal Segment Networks architecture that focused on key video segments to reduce redundancy. MVFNet \cite{wu2021mvfnet} uses multi-view 3D signals to capture video dynamics. Charles et al. \cite{charles2016personalizing}  used frame annotations for posture detection. Kanazawa et al. \cite{kanazawa2019learning} pre-trained model with a temporal encoder for 3D human analysis. Recently, several models can detect poses in real-time. Such models \cite{bazarevsky2020blazepose, cao2017realtime} use deep learning architectures for pose estimation. Yu et al. \cite{yu2018estimating} and Yu et al. \cite{yu2019automatic} focused on action recognition and used CV and insole pressure sensors for ergonomic workload assessment. In contrast, we suggest a technique for quantifying motion levels, assuming comparable action distributions. In this paper, we chose MediaPipe’s pose estimator \cite{lugaresi2019mediapipe}, which uses BlazePose \cite{bazarevsky2020blazepose}.

\subsection{Research Gaps in Worker Motion Analysis}
Worker motion analysis is essential for workplace safety and efficiency. Several studies analyze ergonomic aspects of worker motion, emphasizing worker-friendly workspace design \cite{bortolini2020motion, golabchi2015automated}. Gouett \cite{gouett2010activity} shows the potential of optimizing workflows to improve productivity. Moran \& Wallace \cite{moran2007eccentric} highlight challenges in evaluating worker performance, calling for more robust methods. There is a gap in non-intrusive solutions to preemptively mitigate risks in real-world work conditions \cite{chen2019proactive}. The potential of AI-aided worker training is noted by Ryu et al. \cite{ryu2018motion}, yet a comprehensive framework that integrates CV tools and real-time motion analysis is lacking. The proposed framework address these gaps real-time integration of CV and statistical control techniques, setting a new standard for worker motion monitoring.

\section{Collected and Benchmark Data Description}

\begin{figure*}[h]
  \centering
  \includegraphics[width=\textwidth]{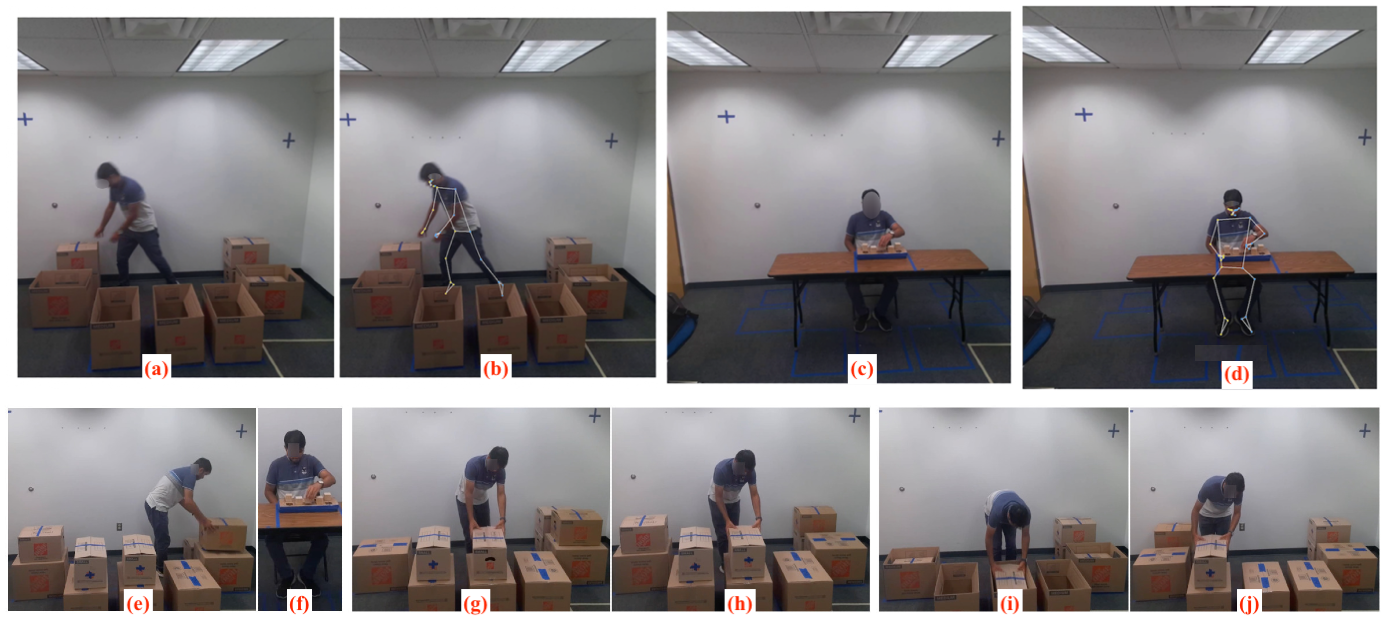}
    \caption{(a) Standing assembly task with feet obscured. (b) Posture estimation with MediaPipe to detect obscured feet. (c) Sitting assembly task with trunk and lower limbs obscured. (d) Posture estimation with MediaPipe to detect obscured trunk and lower limbs. (e) L task, where the participant moves big cardboard boxes. (f) S task, where the participant moves small wooden cubes. (g) G task, where the participant follows an guided order as seen in the specific placement of the boxes. (h) U task, where the participant follows an unguided order as seen in the serial (random) placement of the boxes. (i) I task, where the participant inserts the box or cube inside the cavity. (j) P task, where the participant places the box or cube on the platform.
The tasks involve guided or unguided insertion or placement of boxes and cubes.}
    \label{fig:Figure 2}
\end{figure*}

\subsection{Experiment Data Collected}
The experimental design consisted of three components, resulting in eight tasks per participant (2 × 2 × 2). Participants engaged in tasks with varied-sized objects: large boxes (L) or small wooden cubes (S) (see Fig. 2). They followed either guided (G) with a numbering/alphabet system or unguided (U) with random placements. Tasks involved inserting (I) objects into a cavity or placing (P) them on top of a surface. The codes (L, S, G, U, I, and P) were used in combination (e.g., SGI, LUP; see Fig. 2) to label videos for each variable combination. Video recordings from two healthy males (age 27 ± 1.41 years; height 1.74 ± 0.014 meters; arm length 0.70 ± 0.014 meters) were collected at 30 frames per second (totaling approximately 5,000 frames per participant). The Institutional Review Board (IRB) at Arizona State University approved the experimental protocol (STUDY00016442). Before participating in the experiment, participants read and signed an IRB-approved informed written consent form.

The experiment included guided and unguided techniques to test how explicit guidance affects participants' performance. Guided tasks required structured, sequential instructions based on a numbering system, while unguided tasks involved random, self-directed actions. The space was kept free of obstacles to isolate the effects of object size, technique, and action type. Video cameras were positioned at one frontal 0-degree angle and two side angles at 45 degrees to capture upper and lower body motions. Video data were recorded using three devices: a Canon VIXIA 4K camcorder, a consumer-grade 4K camcorder, and a Samsung Galaxy Android smartphone. The workspace measured approximately 4.2 m × 3.9 m with a height of 2.4 m, providing ample space for participants to perform tasks without obstruction. The total floor area was approximately 16.4 m\textsuperscript{2}. Multiple camera angles minimized blind spots and allowed detailed analysis of task strategies.

Recorded videos were manually labeled by task (e.g., SGI, LUP) based on the experimental protocol, and organized with metadata (participant ID, trial number, camera angle). MediaPipe was applied to extract 3D landmarks from each frame for motion quantification and statistical analysis.

\subsection{Benchmark Datasets}

To assess how well our framework generalizes, we used multiple publicly available benchmark datasets covering diverse motion patterns and task types. The G-AI-HMS dataset \cite{iyer2025gaihms} includes synthetic but high-fidelity human motions generated using a generative AI pipeline and validated through pose estimation. These tasks range from upper-limb activities to whole-body motions. The UCF Sports Action dataset \cite{rodriguez2008action, soomro2015action} comprises authentic sports footage with full-body movements recorded in uncontrolled settings, which presents a challenging benchmark for motion-based classification. The UCF50 dataset \cite{reddy2013recognizing} includes a wide variety of human actions sourced from online videos, representative of high inter-class variability, background noise, and occlusions. The PE-USGC dataset \cite{10742356} presents near-duplicate repetitive culinary tasks. These datasets were selected to assess performance across structured, semi-controlled, and dynamic physical environments. For the in-house experimental dataset, we also collected NASA TLX subjective ratings \cite{hart1988development}, which was a perceptual validation benchmark for the motion-based risk warnings.

\section{Motion Amount Quantification and Data Analysis Methodology}
\subsection{Body Landmark Extraction with CV Tools}
We used MediaPipe \cite{lugaresi2019mediapipe} for CV analysis of landmark extraction from video frames. This process involves identifying key landmark points. These landmarks represent upper and lower limbs joints. Motion is calculated from positional changes between frames \cite{iyeranalyzing} for frame steps (0, 2, 4). We treat the data as a continuous multivariate variate stream from a single distribution. Feature vectors per frame capture spatial landmark data. Data smoothing was omitted to preserve real-time variability and analyze raw, naturalistic motion.

Even though the BlazePose model from MediaPipe was introduced in 2019, it continues to be an appropriate choice for tracking ergonomic movements. This is because of its real-time responsiveness, minimal computational demands, and dependable accuracy across diverse motion types. Recent evaluations further established its reliability. Latreche et al. \cite{latreche2023reliability} reported high agreement for shoulder joint motion, with intra-class correlation coefficients exceeding 0.80 and angular discrepancies as small as –0.6 degrees when compared with standard goniometric and inclinometer readings. Likewise, Hamilton et al. \cite{hamilton2024comparison} demonstrated that MediaPipe offered smooth and precise joint trajectory tracking, showing coefficients of variation below 10\% and strong intra-class consistency with Qualisys 3D motion capture for elbow and knee flexion-extension assessments. These findings establish that MediaPipe continues to be an accurate solution for quantifying movement in workplace contexts.

\begin{table}[ht]
\centering
\caption{Landmark Points for S and L Tasks}
\begin{tabular}{|c|c|c|}
\hline
\textbf{Index} & \textbf{Body Part} & \textbf{Task Type} \\ \hline
11 & Left shoulder & L \\ \hline
12 & Right shoulder & L \\ \hline
13 & Left elbow & S and L \\ \hline
14 & Right elbow & S and L \\ \hline
15 & Left wrist & S \\ \hline
16 & Right wrist & S \\ \hline
17 & Left pinky & S \\ \hline
18 & Right pinky & S \\ \hline
19 & Left index & S \\ \hline
20 & Right index & S \\ \hline
21 & Left thumb & S \\ \hline
22 & Right thumb & S \\ \hline
25 & Left knee & L \\ \hline
26 & Right knee & L \\ \hline
\end{tabular}
\end{table}

The landmark points designated for S and L tasks (see Table II) are key to analyzing task-specific movements. For S tasks, involving precise hand movements with small cubes, landmark points such as the left and right elbows (Indexes 13, 14), wrists (15, 16), and various finger joints including pinkies (17, 18), indexes (19, 20), and thumbs (21, 22) were selected to capture fine motor movements. For L tasks with larger boxes, landmark points like left and right shoulders (11, 12) and knees (25, 26) were selected to help analyze larger, exertive movements. By targeting these specific landmark points, researchers and ergonomists can create interventions to reduce workplace injuries and improve efficiency, for task-specific demands.

To calculate the motion amount, we first define the position vector for each landmark \( i \) in frame \( k \) as shown in equation (1).
\begin{equation}
\boldsymbol{v}_{i,k} = (x_{i,k}, y_{i,k}, z_{i,k})
\end{equation}
where \( \boldsymbol{v}_{i,k} \) represents the 3D position vector of landmark \( i \) at frame \( k \), with \( x_{i,k} \), \( y_{i,k} \), and \( z_{i,k} \) as its Cartesian coordinates.

The motion \( \boldsymbol{m}_{i,k} \) for each landmark is then defined as the vector difference between consecutive frames:
\begin{equation}
\boldsymbol{m}_{i,k} = \boldsymbol{v}_{i,k} - \boldsymbol{v}_{i,k-step}
\end{equation}
where \( \boldsymbol{m}_{i,k} \) is the motion vector of landmark \( i \) from frame \( k-step \) and frame \( k \), and \( \boldsymbol{m}_k = [\boldsymbol{m}_{1,k}, \boldsymbol{m}_{2,k},..., \boldsymbol{m}_{p,k}] \). 

The total motion amount \( M_k \) in frame \( k \) is then computed as the sum of magnitudes of each of the \( p \) landmarks' motion vector calculated in equation (2).
\begin{equation}
M_k = \sum_{i=1}^{p} \| \boldsymbol{m}_{i,k} \|
\end{equation}

This calculation quantifies the overall motion \( M_k \) in each frame by summing the magnitudes of all \( p \) landmark motion vectors, capturing the total activity of all landmarks in the frame sequence.

\subsection{Motion Amount Monitoring and Warning}
In this study, we propose using a control chart for systematic monitoring of worker demand statistics. Specifically, the Hotelling's $T^2$ control chart \cite{montgomery2019introduction, 10752596} tracks key features. An upper control limit (UCL) flags outliers as potential warning levels. We use the Hotelling's $T^2$ statistic for warnings based on the joint motion. Williams et al. \cite{williams2006distribution} showed that using T$^2$ statistics derived from the successive differences estimators improve UCL accuracy for individual observations during a Phase I analysis. This statistic monitors the multivariate distribution of the selected features across video frames.

% In multivariate statistical quality control, a continuous quality characteristic is modeled by a multivariate normal distribution with probability density function, as shown in equation (4).
%\begin{equation}
%    f(\boldsymbol{m}_{k}) = \frac{1}{(2\pi)^{\frac{3p}{2}} |\boldsymbol{\Sigma}|^{\frac{1}{2}}} e^{-\frac{1}{2} (\boldsymbol{m}_{k}-\boldsymbol{\mu})^T \boldsymbol{\Sigma}^{-1} (\boldsymbol{m}_{k}-\boldsymbol{\mu})}
%\end{equation}
%\[ -\infty < \boldsymbol{m}_{k} < \infty \]
%where \( \boldsymbol{\mu} \) is the mean motion vector across all \( p \) landmark points in frame \( k \). \( \boldsymbol{\Sigma} \) is the true covariance matrix, representing the covariance between each pair of landmark points in frame \( k \). \( (2\pi)^{\frac{p}{2}} |\boldsymbol{\Sigma}|^{\frac{1}{2}} \) normalizes the density function.

\subsection{Monitoring Statistic}
Consider the matrix \( \boldsymbol{\mathcal V} \) representing the positions of landmarks in 3D space over \( k \) frames, where each frame contains data for landmark indexes from 13 to 22. The matrix is defined in equation (4).

\begin{equation}
\scalebox{0.8}{
$\boldsymbol{\mathcal V} = \left[
\begin{array}{cccc}
\boldsymbol{v}_{13,1} & \boldsymbol{v}_{14,1} & \cdots & \boldsymbol{v}_{22,1} \\
\boldsymbol{v}_{13,2} & \boldsymbol{v}_{14,2} & \cdots & \boldsymbol{v}_{14,2} \\
\vdots & \vdots & \ddots & \vdots \\
\boldsymbol{v}_{13,K} & \boldsymbol{v}_{14,K} & \cdots & \boldsymbol{v}_{22,K} \\
\end{array}
\right]_{K \times 3p}$
}
\end{equation}
where each entry $\boldsymbol{v}_{i, k}=(x_{i, k}, y_{i, k}, z_{i, k})$ denotes the Cartesian coordinates of landmark $i$ in frame $k$. The matrix dimensions are $K \times 3p$, with $K$ representing the number of frames and $p$ as the total landmarks points (e.g., 13 to 22 in this case).

The sample covariance matrix is defined as:
\begin{equation}
\boldsymbol{S} = \frac{1}{K-1} \sum_{i=1}^K (\boldsymbol{m}_{k} - \overline{\boldsymbol{m}})(\boldsymbol{m}_{k} - \overline{\boldsymbol{m}})^T
\end{equation}

For example, hand movements in frame $k$ use landmarks from \( \boldsymbol{\mathcal V} \) like $\boldsymbol v_{13, k}=(x_{13, k}, y_{13, k}, z_{13, k}), \boldsymbol v_{14, k}=(x_{14, k}, y_{14, k}, z_{14, k}), \ldots$ to calculate aggregate motion vector $\boldsymbol{m}_k$.

Based on equation (5), the monitoring statistic for frame \( k \) is calculated as:
\begin{equation}
t_k = (\boldsymbol{m}_k - \boldsymbol{\overline{m}}) \boldsymbol{S}^{-1} (\boldsymbol{m}_k - \boldsymbol{\overline{m}})^T
\end{equation}

where \( t_k \) is the monitoring statistic value for the frame \( k \) assessing the worker motion's extremeness or anomaly status relative to the distribution.

Calculating a control value based on equation (6) for each instance in frames gives us the following vector:
\begin{equation}
\boldsymbol{T} = \begin{bmatrix}
    t_1 \\
    t_2 \\
    \vdots \\
    t_K
\end{bmatrix}
\end{equation}

It is unnecessary to calculate the mean of the mean vectors since the data is not divided into subgroups. Thus, our control values are derived as above in equation (7). The UCL and the lower control limits (LCL) are given in equations (8) and (9) respectively \cite{nistcontrolchart, montgomery2019introduction}. \( F \) is a critical value from the F-distribution (shown in Appendix V of \cite{montgomery2019introduction}) at a significance level \( \alpha \), with the degrees of freedom  \( 3p \) (numerator) and  \( K-3p \) (denominator).

\begin{equation}
\text{UCL} = \frac{3p (K+1)(K-1)}{K^2-3Kp} F_{1-\alpha/2,3p,K-3p}
\end{equation}

\begin{equation}
\text{LCL} = 0
\end{equation}

\subsection{Control Limits}
UCL and LCL are identify significant variation. The LCL shows if the process meets standards or signals decreased variation. The LCL represents motion quality. Control chart techniques have two phases: Phase I (offline) and Phase II (online) monitoring. A video is split for offline training (establishing baseline patterns) and online monitoring (real-time worker demand evaluation). The LCL is defined as zero, describing the minimum possible variation in worker motion. In this context, it acts as a baseline threshold to identify frames with no significant motion activity. While LCLs are typically used in quality control, here they indicate periods of stillness or negligible motion.

\subsection{Correlation Analysis to Validate Significant Motions}
To assess motion variation, we compute the Root Mean Square Deviation (RMSD) \cite{wada2017root}, focusing on tasks involving smaller objects (S tasks), which typically exhibit intricate, frequent movements. RMSD is calculated using equation (10).
\begin{equation}
\text{RMSD}_{\text{task}} = \sqrt{\frac{1}{K} \sum_{k=1}^{K} (M_k - \overline{M})^2}
\end{equation}
where \( \overline{M} \) is the mean motion amount across all frames, calculated similar to equation (3). The full pipeline for motion quantification and adaptive ergonomic risk classification is summarized in Algorithm 1.

\subsection{Random Forest-Based Classification for Predicting Ergonomic Warnings}

To assess whether motion-derived features can generalize ergonomic risk classification across tasks, we used a supervised learning pipeline with Random Forests. The objective was to predict whether a task video segment should be flagged as high-risk, using motion statistics and control chart outputs. Hotelling’s \( T^2 \) control chart finds frame-level motion issues but needs task-specific setup and may not scale well across different task types. To address this, we used the warning outputs from the \( T^2 \) control chart as a basis for training a classifier that could learn broader patterns of motion-related risk.

\subsubsection{Label Assignment Using Adaptive Thresholding}

Binary warning labels were generated using adaptive thresholding that dynamically adjusts to the baseline activity of each dataset. For each dataset \( d \), a threshold \( \tau_d \) was computed as the median number of motion-based warnings. Each video segment \( i \) was assigned a binary label \( y_i \), indicating whether the number of warnings exceeded the threshold, as defined in equation (11). This adaptive method balances each dataset without relying on a global threshold, which could be unsuitable due to variable baseline motion patterns across datasets.

\begin{equation}
y_i = 
\begin{cases}
1 & \text{if } \text{warnings}_i > \tau_d \\
0 & \text{otherwise}
\end{cases}
\tag{11}
\end{equation}

\subsubsection{Feature Construction}

The classifier input included four features per video segment: RMSD of joint motion, mean motion amount, standard deviation of motion amount, and the Hotelling-based upper control limit (UCL). These features capture variability in motion behavior and anomaly thresholds derived from multivariate monitoring. Each input sample \( \mathbf{x}_i \in \mathbb{R}^4 \) representing the task segments' motion profile was constructed as shown in equation (12).

\begin{equation}
\mathbf{x}_i = \left[
\text{RMSD}_i,\ 
\mu_{\text{motion},i},\ 
\sigma_{\text{motion},i},\ 
\text{UCL}_i
\right]
\tag{12}
\end{equation}

\begin{algorithm}[ht]
\caption{Motion Quantification and Adaptive Risk Classification Pipeline}
\begin{algorithmic}[1]
\STATE \textbf{Input:} Raw task video directory \( \mathcal{V} \), list of landmark sets \( \mathcal{L} \), step sizes \( \mathcal{S} \)
\STATE \textbf{Output:} Labeled dataset with warnings, classifier performance metrics

\STATE \textbf{Stage 1: Landmark Extraction via MediaPipe}
\FOR{each video folder \( v \in \mathcal{V} \)}
    \FOR{each frame \( f \in v \)}
        \STATE Read frame and convert to RGB
        \STATE Extract 3D landmarks using MediaPipe: \( \mathbf{v}_{i,k} = (x_{i,k}, y_{i,k}, z_{i,k}) \)
    \ENDFOR
    \STATE Save per-frame landmark matrix to CSV
\ENDFOR

\STATE \textbf{Stage 2: Motion Quantification Across Frames}
\FOR{each CSV file}
    \FOR{each step size \( s \in \mathcal{S} \)}
        \FOR{each landmark set \( \ell \in \mathcal{L} \)}
            \STATE Compute motion vectors: \( \boldsymbol{m}_{i,k} = \boldsymbol{v}_{i,k} - \boldsymbol{v}_{i,k-s} \)
            \STATE Compute total motion: \( M_k = \sum_{i=1}^p \| \boldsymbol{m}_{i,k} \| \)
            \STATE Compute Hotelling's \( T^2 \) statistic for each frame: \( t_k \) \COMMENT{Eq. (6)}
            \STATE Estimate control limit: \( \text{UCL} \) from F-distribution \COMMENT{Eq. (8)}
            \STATE Count warnings: \( w = \sum_k [t_k > \text{UCL}] \)
            \STATE Calculate RMSD of motion: \( \text{RMSD} = \sqrt{\frac{1}{K} \sum_k (M_k - \bar{M})^2} \)
            \STATE Append metrics to results table
        \ENDFOR
    \ENDFOR
\ENDFOR

\STATE \textbf{Stage 3: Adaptive Warning Labeling}
\FOR{each dataset \( D \)}
    \STATE Compute median warning value: \( \theta_D = \text{median}(w_D) \)
    \STATE Label instances: \( y_k = \mathbb{1}(w_k > \theta_D) \)
\ENDFOR

\STATE \textbf{Stage 4: Classifier Training and Evaluation}
\STATE Select source dataset \( D_{\text{train}} = \texttt{Assembly} \)
\STATE Train Random Forest on \( D_{\text{train}} \): features = [RMSD, mean motion, std motion, UCL]
\FOR{each dataset \( D_{\text{test}} \)}
    \STATE Predict warning labels using trained model
    \STATE Compute classification metrics: precision, recall, accuracy, F1
\ENDFOR

\STATE \textbf{return} Labeled results, classification performance metrics
\end{algorithmic}
\end{algorithm}

\subsubsection{Model Training and Inference}

A Random Forest model \( f_\theta \) was trained using the Assembly dataset. The model learns to associate the statistical features in \( \mathbf{x}_i \) with the warning labels \( y_i \) derived from the \( T^2 \) control chart. For a new segment, the predicted label \( \hat{y}_i \) is derived from the trained model, as shown in equation (13). The output \( \hat{y}_i \in \{0, 1\} \) indicates whether the input motion segment is predicted to be low-risk (0) or high-risk (1).

\begin{equation}
\hat{y}_i = f_{\theta}(\mathbf{x}_i)
\tag{13}
\end{equation}

The Random Forest aggregates predictions from an ensemble of \( T \) decision trees. Each tree \( h_j \) makes its own prediction, and the final result is based on majority vote, as shown in equation (14). This ensemble approach prevents overfitting and across data domains.

\begin{equation}
f_{\theta}(\mathbf{x}) = \text{mode}\left( \{ h_j(\mathbf{x}) \}_{j=1}^T \right)
\tag{14}
\end{equation}

This classification layer provides real-time decision that extends risk evaluation beyond frame-level limits that can be used across tasks.

\subsubsection{Cross-Dataset Evaluation}

After training on the Assembly dataset, the model was evaluated on all other datasets using the adaptive thresholding rule for labeling. Performance was quantified using standard classification metrics—precision, recall, F1-score, and accuracy.

Precision is the proportion of correctly identified positive instances out of all predicted positives, shown in equation (15).

\begin{equation}
\text{Precision} = \frac{TP}{TP + FP}
\tag{15}
\end{equation}

Recall, the proportion of correctly identified positives among all actual positives, is given in equation (16).

\begin{equation}
\text{Recall} = \frac{TP}{TP + FN}
\tag{16}
\end{equation}

The F1 score is the balance between precision and recall using their harmonic mean, calculated using equation (17).

\begin{equation}
\text{F1-score} = \frac{2 \cdot \text{Precision} \cdot \text{Recall}}{\text{Precision} + \text{Recall}}
\tag{17}
\end{equation}

Finally, accuracy, the proportion of correctly classified instances over all instances, is shown in equation (18).

\begin{equation}
\text{Accuracy} = \frac{TP + TN}{TP + TN + FP + FN}
\tag{18}
\end{equation}

Here, \( TP \), \( FP \), \( TN \), and \( FN \) represent true positives, false positives, true negatives, and false negatives respectively. These metrics quantify classifier performance across domains and conditions of class imbalance.

\section{Results}
\subsection{Assembly Task Analysis}
The results reveal trends in motion amount, velocity, acceleration, landmark points, and control charts. All standard deviation (SD) data in this manuscript refer to population SD. For motion amount in Table III, the mean decreases over steps, while velocity and variability are low. Acceleration exhibits high initial variability (SD of 0.311 at Step 0), which decreases by Step 4. In Table IV, landmark point movement data indicate that smaller object tasks (S) have finer X/Y mean variation, whereas larger object tasks (L) show greater Z-axis variability, particularly for tasks LGI and LGP. Table IV highlights that warnings are more frequent for L tasks than S tasks, with 75 and 77 warnings for  LGI and LGP, respectively, indicating greater variability. Larger object tasks exhibit higher variability in landmark points and warnings, while motion amount and velocity data suggest greater movement stability. For S tasks, the RMSD (described in equation (12) for calculated aggregate joint motion amount) was found to be 0.072 meters, greater than the 0.035 meters observed for L tasks. Additionally, the Pearson correlation coefficient for motion amount and the Hotelling's $T^2$ statistic was approximately 35\% higher for S tasks than for L tasks \cite{cohen2009pearson}.

\begin{table}[!htbp]
\caption{Statistics for Joint Kinematics at Different Steps (Motion, Velocity, and Acceleration computed in normalized coordinate space; unitless)}
\label{tab:combined_stats}
\centering
\begin{tabularx}{\columnwidth}{|>{\centering\arraybackslash}c|>{\centering\arraybackslash}c|>{\centering\arraybackslash}X|>{\centering\arraybackslash}X|>{\centering\arraybackslash}X|}
\hline
\textbf{Data Type} & \textbf{Statistic} & \textbf{Step 0} & \textbf{Step 2} & \textbf{Step 4} \\ \hline
\multirow{3}{*}{Motion Amount} 
& Count & 1,099 & 1,097 & 1,095 \\ \cline{2-5} 
& Mean & 0.097 & 0.052 & 0.057 \\ \cline{2-5} 
& SD & 0.086 & 0.068 & 0.072 \\ \cline{2-5} \hline

\multirow{3}{*}{Velocity} 
& Count & 1,099 & 1,097 & 1,095 \\ \cline{2-5} 
& Mean & 0.012 & 0.003 & 0.002 \\ \cline{2-5} 
& SD & 0.01 & 0.004 & 0.002 \\ \cline{2-5} \hline

\multirow{3}{*}{Acceleration} 
& Count & 1,098 & 1,096 & 1,094 \\ \cline{2-5} 
& Mean & 0.0006 & 0.00004 & -0.00001 \\ \cline{2-5} 
& SD & 0.311 & 0.078 & 0.02 \\ \cline{2-5} \hline

\end{tabularx}
\end{table}

\begin{table}[htbp]
\centering
\caption{Statistics for Landmark Point Movement}

\begin{tabularx}{\columnwidth}{|>{\centering\arraybackslash}X|>{\centering\arraybackslash}X|>{\centering\arraybackslash}X|>{\centering\arraybackslash}X|>{\centering\arraybackslash}X|>{\centering\arraybackslash}X|>{\centering\arraybackslash}X|}

    \hline
    \multirow{2}{*}{\textbf{Task}} & \multicolumn{3}{c|}{\textbf{Mean}} & \multicolumn{3}{c|}{\textbf{SD}} \\ \cline{2-7}
                                   & \textbf{X} & \textbf{Y} & \textbf{Z} & \textbf{X} & \textbf{Y} & \textbf{Z} \\ \hline
    SGI & 0.604 & 0.546 & 0.011 & 0.018 & 0.021 & 0.035 \\ \hline
    SGP & 0.611 & 0.527 & 0.023 & 0.011 & 0.020 & 0.047 \\ \hline
    SUI & 0.609 & 0.523 & 0.015 & 0.012 & 0.022 & 0.050 \\ \hline
    SUP & 0.607 & 0.530 & 0.020 & 0.013 & 0.019 & 0.045 \\ \hline
    LGI & 0.534 & 0.378 & -0.134 & 0.080 & 0.055 & 0.140 \\ \hline
    LGP & 0.537 & 0.385 & -0.120 & 0.085 & 0.065 & 0.150 \\ \hline
    LUI & 0.532 & 0.370 & -0.125 & 0.075 & 0.050 & 0.130 \\ \hline
    LUP & 0.529 & 0.365 & -0.115 & 0.070 & 0.045 & 0.120 \\ \hline
\end{tabularx}
\end{table}

\begin{table}[h]
\centering
\caption{Correlation between Motion Warnings and NASA-TLX Subscales}
\begin{tabular}{|l|c|c|}
\hline
\textbf{NASA-TLX Subscale} & \textbf{Correlation (r)} & \textbf{p-value} \\ \hline
Mental Demand     & 0.192 & $p < 0.01$ \\ \hline
Physical Demand   & 0.214 & $p < 0.005$ \\ \hline
Temporal Demand   & 0.231 & $p < 0.005$ \\ \hline
Performance       & 0.109 & $p = 0.13$ \\ \hline
Effort            & 0.155 & $p < 0.05$ \\ \hline
Frustration       & ---   & ---        \\ \hline
\textbf{Average}  & \textbf{0.218} & \textbf{$p < 0.005$} \\ \hline
\end{tabular}
\label{tab:nasa_tlx_corr}
\end{table}

Table V shows the Pearson correlations between motion warnings generated by the framework and subjective workload ratings from the NASA-TLX subscales. Statistically significant positive correlations were observed for Mental Demand (r = 0.192, p $<$ 0.01), Physical Demand (r = 0.214, p $<$ 0.005), Temporal Demand (r = 0.231, p $<$ 0.005), and Effort (r = 0.155, p $<$ 0.05), indicating that increased motion warnings tend to co-occur with higher perceived workload in these domains. Performance showed a weak, non-significant correlation (r = 0.109, p = 0.13), while Frustration was excluded due to incomplete data. The average correlation across subscales was 0.218 (p $<$ 0.005), suggesting a meaningful alignment between objective motion anomalies and subjective perceptions of workload. 

\subsection{Benchmarking with Body-part Ablation}
Table VI presents the mean and standard deviation for RMSD, mean motion, standard motion, and warning counts across all datasets and ablation types, averaged over step sizes and tasks.

Across all datasets, the full body ablation consistently showed the highest RMSD and motion metrics. In the Assembly dataset, full body yielded an RMSD of 0.670 ± 0.313 and a mean motion of 0.829 ± 0.462, while hands and lower body subsets resulted in notably lower values. A similar pattern was observed in G-AI-HMS, where full body RMSD was 0.761 ± 0.291 compared to 0.229 ± 0.061 for the hands subset.

In the PE-USGC (Near-duplicate) dataset, full body values were substantially higher (RMSD: 3.008 ± 1.598; mean motion: 4.126 ± 1.740) than the partial ablations. UCF Sports Action showed elevated motion values (e.g., full body mean motion: 4.479 ± 2.727) but low warning counts across all ablations. Notably, UCF50 recorded the highest overall RMSD and motion values, with full body RMSD reaching 4.184 ± 3.098 and mean motion at 6.942 ± 4.303. However, warning values remained low across datasets except for Assembly, which had the highest warnings under full body (195.292 ± 64.307).

Partial ablations (hands, lower body, upper body) consistently reduced all motion metrics and warnings relative to full body inputs. This trend was stable across the five datasets, with low variation in standard deviations between corresponding types of ablation.

\noindent
Table VII reports classification performance across five datasets using adaptive thresholding for warning detection. Performance is evaluated using recall, accuracy, precision, and F1-score.

The model trained on the Assembly dataset achieved the highest and most balanced performance across all metrics, with recall, accuracy, precision, and F1-score all equal to 0.94, as shown in Table VII. This shows that the classifier generalizes well within the source domain and has a low false positive rate. The alignment of precision and recall suggests that the classifier is not sensitive to warning conditions in this dataset.

For the G-AI-HMS dataset, the model retained performance with scores around 0.80 across all metrics. The slightly higher precision (0.81) compared to recall (0.80) implies that the model was conservative in issuing warnings, successfully avoiding many false alarms while still capturing most true warning instances. Performance decreased on the PE-USGC dataset, with recall dropping to 0.66 and F1-score to 0.63, despite relatively high precision (0.74). This shows that the model accurately identifies warnings when it predicts them, but misses many actual warning cases, suggesting under-detection due to domain differences from the Assembly training data.

In the UCF Sports Action dataset, the model achieved a moderate recall of 0.71 but exhibited low precision (0.57) and F1-score (0.53). These results suggest over-prediction of warnings, potentially due to the dynamic nature of sports-related high-risk activity. Similarly, UCF50 produced similar trends with balanced recall and accuracy (0.65) and slightly higher precision (0.68). The F1-score of 0.64 indicates consistent, if modest, classification ability. The large and heterogeneous nature of this dataset likely contributes to its stable but non-optimal performance.

Adaptive thresholding showed the strongest results on datasets structurally similar to the training source (Assembly and G-AI-HMS). Performance dropped on more diverse or high-motion datasets, highlighting the difficulty of using fixed model settings across physical tasks.

\begin{table*}[ht]
\centering
\scriptsize
\caption{Descriptive statistics by dataset and ablation type (averaged over step sizes and tasks).}
\label{tab:descriptive_stats_tabular}
\resizebox{\textwidth}{!}{
\begin{tabular}{|l|l|c|c|c|c|}
\hline
\textbf{Dataset} & \textbf{Ablation} & \textbf{RMSD} & \textbf{Motion Mean} & \textbf{Motion SD} & \textbf{Warnings} \\
\hline
\multirow{4}{*}{G-AI-HMS \cite{iyer2025gaihms}} 
 & Full Body   & 0.761 ± 0.291 & 0.950 ± 0.297 & 0.761 ± 0.291 & 103.375 ± 114.671 \\ \cline{2-6}
 & Hands       & 0.229 ± 0.061 & 0.297 ± 0.107 & 0.229 ± 0.061 & 105.333 ± 91.716 \\ \cline{2-6}
 & Lower Body  & 0.307 ± 0.152 & 0.304 ± 0.118 & 0.307 ± 0.152 & 107.208 ± 101.138 \\ \cline{2-6}
 & Upper Body  & 0.293 ± 0.076 & 0.386 ± 0.128 & 0.293 ± 0.076 & 109.875 ± 103.227 \\ \cline{2-6}
\hline
\multirow{4}{*}{PE-USGC \cite{10742356}} 
 & Full Body   & 3.008 ± 1.598 & 4.126 ± 1.740 & 3.008 ± 1.598 & 114.958 ± 90.720 \\ \cline{2-6}
 & Hands       & 0.667 ± 0.357 & 0.864 ± 0.469 & 0.667 ± 0.357 & 108.991 ± 64.320 \\ \cline{2-6}
 & Lower Body  & 1.189 ± 0.570 & 1.678 ± 0.608 & 1.189 ± 0.570 & 113.852 ± 69.345 \\ \cline{2-6}
 & Upper Body  & 0.951 ± 0.513 & 1.231 ± 0.648 & 0.951 ± 0.513 & 113.620 ± 71.310 \\ \cline{2-6}
\hline
\multirow{4}{*}{UCF Sports Action \cite{rodriguez2008action,soomro2015action}} 
 & Full Body   & 2.434 ± 1.656 & 4.479 ± 2.727 & 2.434 ± 1.656 & 0.000 ± 0.000 \\ \cline{2-6}
 & Hands       & 0.757 ± 0.435 & 1.221 ± 0.727 & 0.757 ± 0.435 & 0.766 ± 2.794 \\ \cline{2-6}
 & Lower Body  & 0.767 ± 0.496 & 1.368 ± 0.913 & 0.767 ± 0.496 & 0.286 ± 1.362 \\ \cline{2-6}
 & Upper Body  & 1.013 ± 0.624 & 1.658 ± 0.988 & 1.013 ± 0.624 & 0.234 ± 1.366 \\ \cline{2-6}
\hline
\multirow{4}{*}{UCF50 \cite{reddy2013recognizing}} 
 & Full Body   & 4.184 ± 3.098 & 6.942 ± 4.303 & 4.184 ± 3.098 & 14.743 ± 54.658 \\ \cline{2-6}
 & Hands       & 0.962 ± 0.708 & 1.408 ± 0.976 & 0.962 ± 0.708 & 13.122 ± 47.737 \\ \cline{2-6}
 & Lower Body  & 1.289 ± 1.010 & 2.029 ± 1.487 & 1.289 ± 1.010 & 13.923 ± 49.991 \\ \cline{2-6}
 & Upper Body  & 1.604 ± 1.200 & 2.473 ± 1.761 & 1.604 ± 1.200 & 13.672 ± 48.764 \\ \cline{2-6}
\hline
\multirow{4}{*}{Assembly (current study)} 
 & Full Body   & 0.670 ± 0.313 & 0.829 ± 0.462 & 0.670 ± 0.313 & 195.292 ± 64.307 \\ \cline{2-6}
 & Hands       & 0.252 ± 0.103 & 0.250 ± 0.119 & 0.252 ± 0.103 & 154.833 ± 41.731 \\ \cline{2-6}
 & Lower Body  & 0.210 ± 0.165 & 0.233 ± 0.181 & 0.210 ± 0.165 & 161.500 ± 42.953 \\ \cline{2-6}
 & Upper Body  & 0.315 ± 0.127 & 0.331 ± 0.161 & 0.315 ± 0.127 & 174.146 ± 45.950 \\ \cline{2-6}
\hline
\end{tabular}
}
\end{table*}

\begin{table}[ht]
\centering
\caption{Classification performance (Recall, Accuracy, Precision, and F1-score) using adaptive thresholding.}
\label{tab:adaptive_threshold_results}
\begin{tabular}{|l|c|c|c|c|}
\hline
\textbf{Dataset} & \textbf{Recall} & \textbf{Acc.} & \textbf{Precision} & \textbf{F1} \\
\hline
G-AI-HMS \cite{iyer2025gaihms}              & 0.80 & 0.80 & 0.81 & 0.80 \\ \hline
PE-USGC \cite{10742356}        & 0.66 & 0.66 & 0.74 & 0.63 \\ \hline
UCF Sports Action \cite{rodriguez2008action,soomro2015action}     & 0.71 & 0.65 & 0.57 & 0.53 \\ \hline
UCF50 \cite{reddy2013recognizing}                 & 0.65 & 0.65 & 0.68 & 0.64 \\
\hline
Assembly (current study)              & 0.94 & 0.94 & 0.94 & 0.94 \\ \hline
\end{tabular}
\end{table}

\begin{figure*}[htbp!]
  \centering
  \includegraphics[width=\textwidth, height=300pt]{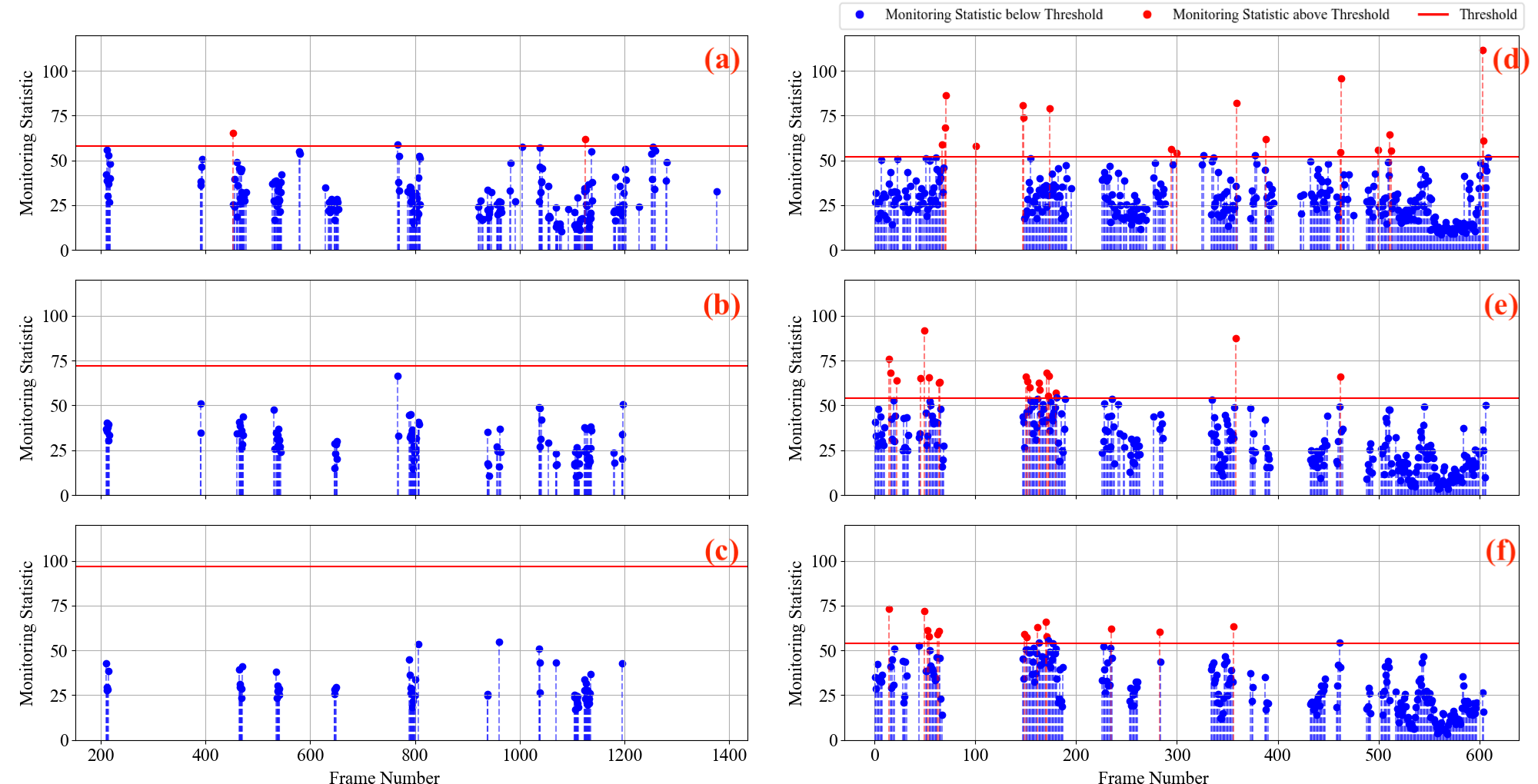}
    \caption{Control charts for SGI: (a) Step 0 (b) Step 2 (c) Step 4 and SGP: (d) Step 0 (e) Step 2 (f) Step 4.}
    \label{fig:Figure 3}
\end{figure*}

\begin{figure*}[htbp!]
  \centering
  \includegraphics[width=\textwidth, height=300pt]{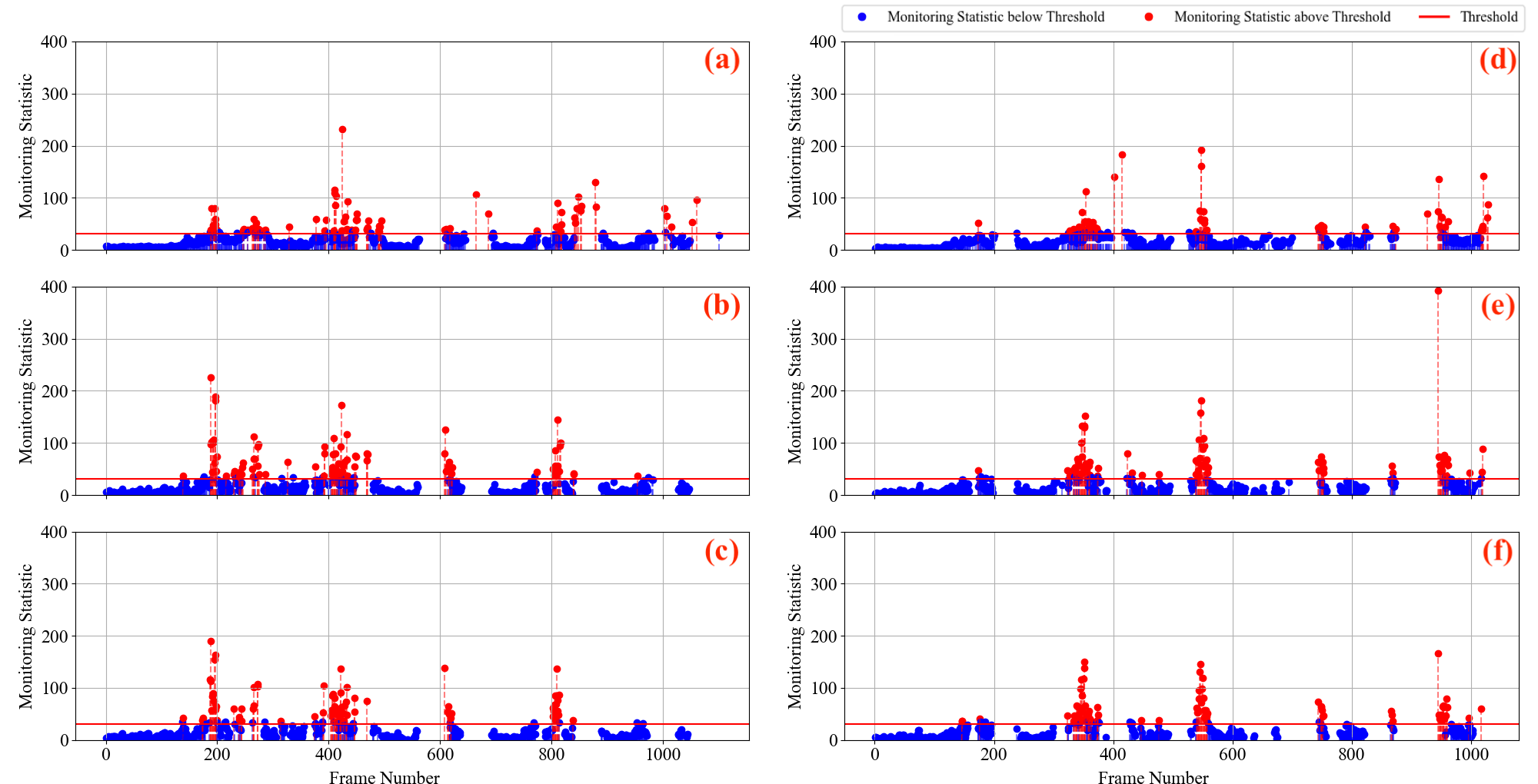}
    \caption{Control charts for LGI: (a) Step 0 (b) Step 2 (c) Step 4 and LGP: (d) Step 0 (e) Step 2 (f) Step 4.}
    \label{fig:Figure 4}
\end{figure*}

\section{Discussion}
The limited camera clarity in S tasks may have caused the CV algorithm’s accuracy to detect postural changes. Perceived motion demand (0-100) scale showed averages of 69.9 for L and 21.9 for S tasks, with data showing larger motion in L task trajectories. LGI landmarks (11, 12, 13, 14, 25, 26) deviated more than SGI’s (13-22). This observation aligns with findings from the study by Zhou et al. \cite{zhou2018raising} that used CV to assess student posture, highlighting the impact of camera resolution on detection accuracy.

Small task (S) trajectories highlight hand movement smoothness; L tasks show needed coordination. S task trajectories detail landmark paths for placement and stabilization. L task trajectories show arm reach and bending, highlighting ergonomic needs. Variability in trajectories suggests technique differences, requiring training or ergonomic intervention. These results agree with Ding et al. \cite{ding2019real}, who created a real-time webcam approach to monitor upper-body postures and similarly highlighted the importance of ergonomic adjustments to address posture variability.

Velocity and motion amount metrics show task dynamics. At Step 0, mean velocity is 0.012 (SD=0.01), peaks at 0.11, dropping to 0.002 (SD=0.002) by Step 4. Initial motion amount is 0.1 (SD=0.086), lowers at Steps 2 and 4 as frame-skipping smoothes fluctuations. Acceleration variability starts high at Step 0 (SD=0.311) and declines, showing controlled motion. This pattern aligns with the study by Chen et al. \cite{chen2023cognitive}, which discusses that increased task complexity can lead to higher variability in human performance and behavior.

As shown in Fig. 3, small tasks have higher mean control chart values (29.55 for SGI, 29.92 for SGP) than large tasks (17.97 for LGP, 17.94 for LUP), likely due to precise, high-activity requirements. Large tasks like LGI and LGP (see Fig. 4) show high motion peaks (392.99 for LGP, 231.44 for LGI) and increased warnings (39 for LUI, 30 for LUP). These high warnings for L tasks indicate frequent deviation from safe movement, needing work optimization or ergonomic intervention. This observation aligns with research on task complexity and physical strain, showing a significant relation between job complexity and increased employee strain \cite{li2017intra}.

Landmark point statistics (see Table IV) differentiate tasks by size, guidance, and action. Small, guided tasks (SGP, SGI) center movements near the median workspace (SGP: X=0.611, Y=0.527; SGI: X=0.604, Y=0.546), while large tasks (LGI, LGP) show wider spread (LGI: X=0.534, Y=0.378). Unguided large tasks (LUI, LUP) have high SDs across coordinates, indicating broader movements due to task demand. Ranges for large (L), unguided (U) tasks (e.g., LGI, LUI) reflect greater spatial demands than small (S), guided (G) tasks, where movement is within controlled spatial limits. These patterns inform tool and workspace designs, guiding objects closer for small, precise tasks and offering support for large, varied tasks. This is in line with findings from the study by Liu \& Li \cite{liu2011toward} on task complexity and task performance, which highlights the importance of workspace design in managing task demands. Compared to traditional methods like REBA \cite{HIGNETT2000201} and RULA \cite{MCATAMNEY199391}, this CV-based framework enables continuous tracking and dynamic alerts, aligning with prior work on real-time monitoring in repetitive tasks.

\section{Conclusion and Future Work}

The findings of this study provide valuable support for ergonomic practitioners and workplace designers \cite{seim2014ergonomics}. By analyzing tasks, professionals can identify high-risk movements and develop workplace-specific interventions. For example, if a task involves frequent, risky reaching movements \cite{d2013control} (detected via z-coordinate analysis), mechanical aids or workspace adjustments can be implemented to reduce such reaches. Simulating task and layout changes in virtual environments offers ergonomic insights before real-world modifications are made \cite{10536400}. Long-term motion tracking enables monitoring the effectiveness of ergonomic solutions, facilitating continuous assessment of workspace changes and strategy adjustments. Motion analysis also helps predict and prevent future issues by minimizing estimation errors \cite{10124779}. The CV-based joint motion amount assessment effectively identifies perceived task demands, with consistent landmark trajectories validating its reliability. However, this study has several limitations, including its retrospective analysis, fixed video angles, and a small sample size (limited to two similar male participants). Additionally, the lack of motion capture validation is another limitation. Future research will focus on designing scalable human subject experiments to further evaluate CV-based methods. In conclusion, the findings underscore the capability of the CV-based technique for providing reliable motion representation. The full implementation code and supporting datasets are publicly available on GitHub \cite{iyer2024cvworkermotion}.

\section*{Acknowledgments}
We thank Boyang Xu for contributions at the project’s outset and Bryan Havens for early data collection assistance. N.M. acknowledges support from ASU’s Master’s Opportunity for Research in Engineering (MORE) program. We are also grateful to all participants in the experiments.

\bibliographystyle{IEEEtran}
\bibliography{template}

\begin{comment}
\newpage

\section{Biography Section}
If you have an EPS/PDF photo (graphicx package needed), extra braces are
 needed around the contents of the optional argument to biography to prevent
 the LaTeX parser from getting confused when it sees the complicated
 $\backslash${\tt{includegraphics}} command within an optional argument. (You can create
 your own custom macro containing the $\backslash${\tt{includegraphics}} command to make things
 simpler here.)
 
\vspace{11pt}

\bf{If you include a photo:}\vspace{-33pt}
\begin{IEEEbiography}[{\includegraphics[width=1in,height=1.25in,clip,keepaspectratio]{fig1}}]{Michael Shell}
Use $\backslash${\tt{begin\{IEEEbiography\}}} and then for the 1st argument use $\backslash${\tt{includegraphics}} to declare and link the author photo.
Use the author name as the 3rd argument followed by the biography text.
\end{IEEEbiography}

\vspace{11pt}

\bf{If you will not include a photo:}\vspace{-33pt}
\begin{IEEEbiographynophoto}{John Doe}
Use $\backslash${\tt{begin\{IEEEbiographynophoto\}}} and the author name as the argument followed by the biography text.
\end{IEEEbiographynophoto}
\end{comment}

\vfill

\end{document}